\documentclass[sigconf]{acmart}

\usepackage{booktabs}
\usepackage{tabularx}
\usepackage{rotating}
\usepackage{multirow}
\usepackage{array}
\usepackage{colortbl}
\usepackage{ulem}
\usepackage{amsmath}

\setcopyright{rightsretained}

\copyrightyear{2017}
\acmYear{2017}
\setcopyright{acmcopyright}
\acmConference{GECCO '17}{July 15-19, 2017}{Berlin,
Germany}\acmPrice{15.00}\acmDOI{http://dx.doi.org/10.1145/3071178.3071296}
\acmISBN{978-1-4503-4920-8/17/07}

\begin{document}
\title{A Minimal Developmental Model Can Increase Evolvability in Soft Robots}


\author{Sam Kriegman}
\affiliation{%
  \institution{Morphology, Evolution \& Cognition Lab \\ University of Vermont}
  \city{Burlington, VT} 
  \country{USA}
}
\email{sam.kriegman@uvm.edu}

\author{Nick Cheney}
\affiliation{%
  \institution{Creative Machines Lab \\ Cornell University}
  \city{Ithaca, NY} 
  \country{USA}
}

\author{Francesco Corucci}
\affiliation{%
  \institution{The BioRobotics Institute \\ Scuola Superiore Sant'Anna}
  \city{Pisa} 
  \country{Italy}
}

\author{Josh C. Bongard}
\affiliation{
  \institution{Morphology, Evolution \& Cognition Lab \\ University of Vermont}
  \city{Burlington, VT} 
  \country{USA}
}

\renewcommand{\shortauthors}{Kriegman et al.}

\begin{abstract}

Different subsystems of organisms adapt over many time scales,
such as rapid changes in the nervous system (learning),
slower morphological and neurological change over the lifetime 
of the organism (postnatal development), and change over
many generations (evolution). Much work has
focused on instantiating learning or evolution in robots, but relatively
little on development. Although many theories have been
forwarded as to how development can aid evolution, it is difficult
to isolate each such proposed mechanism. Thus, here we introduce
a minimal yet embodied model of development: the body of the
robot changes over its lifetime, yet growth is not influenced
by the environment. We show that even this simple developmental
model confers
evolvability because it allows evolution to sweep over a larger
range of body plans than an equivalent non-developmental system,
and subsequent heterochronic mutations
`lock in' this body plan in more morphologically-static descendants.
Future work will involve gradually complexifying the developmental
model to determine when and how such added complexity increases
evolvability.

\end{abstract}



\maketitle

\section{Introduction}
\label{sec:introduction}

\begin{figure*}
\includegraphics[width=1\textwidth]{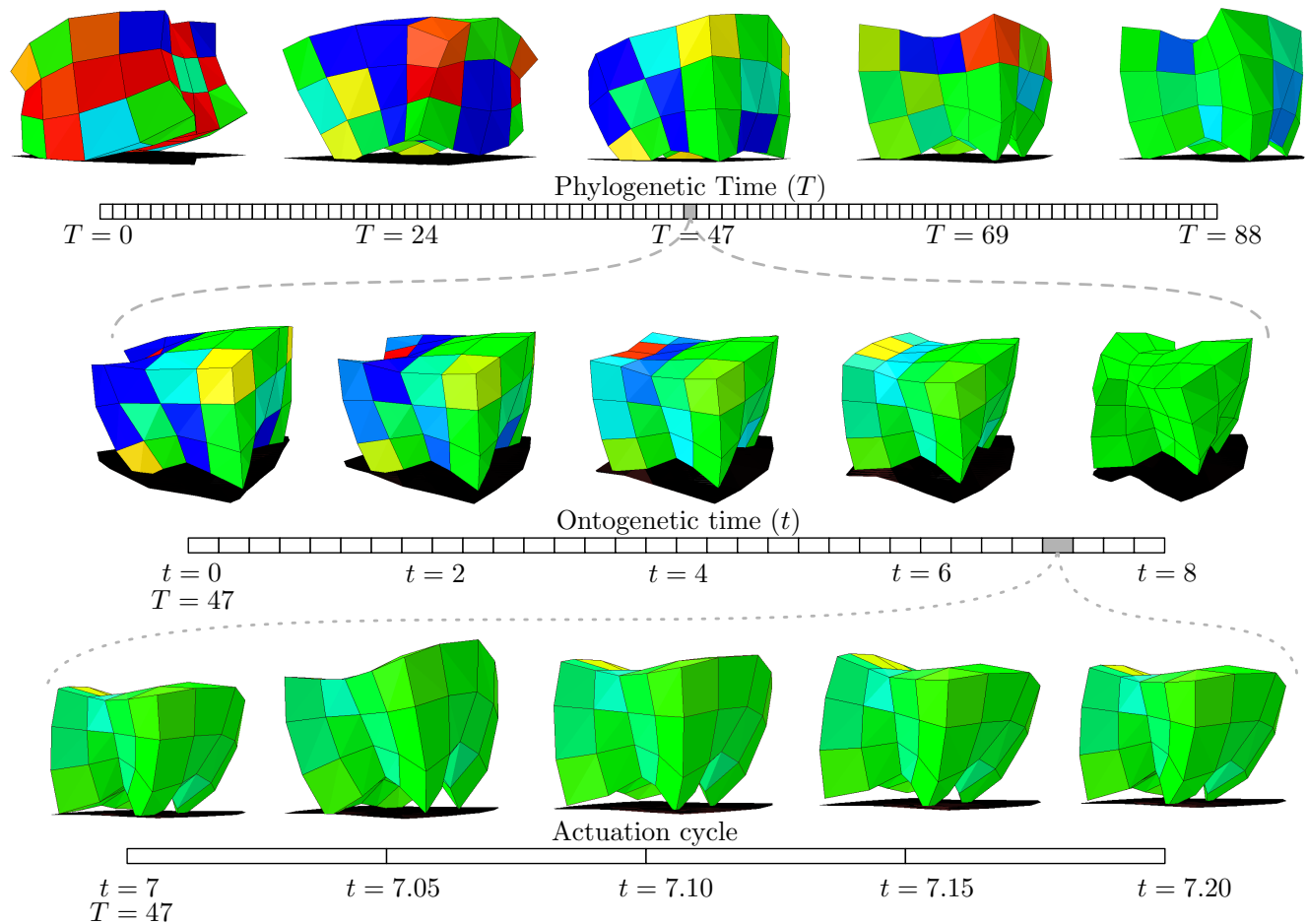}
\vspace{-0.5cm}
\caption{\label{fig:videos}   The evolutionary history of an Evo-Devo robot. One of the five phylogenies is broken down into five ontogenies which is in turn shown at five points in its actuation cycle. Voxel color indicates the remaining development. Blue for shrinkage, red for growing, and green for no further change. This robot is featured in video at \href{https://youtu.be/gXf2Chu4L9A}{https://youtu.be/gXf2Chu4L9A}.}
\end{figure*}

Many theories have been proposed as to how development can confer evolvability.
Selfish gene theory \cite{dawkins1982extended} suggests that prenatal development from a single-celled egg is not a superfluous byproduct of evolution, but is instead a critical process that ensures uniformity among genes contained within a single organism and in turn their cooperation towards mutual reproduction. 
Developmental plasticity, the ability of an organism to modify its form in response to environmental conditions, is believed to play a crucial role in the origin and diversification of novel traits \cite{Moczekrspb20110971}.
Others have shown that development can in effect `encode', and thus
avoid on a much shorter time scale, constraints that would otherwise
be encountered and suffered by non-developmental systems
\cite{kouvaris2017evolution}.

Several models that specifically address development of embodied agents have been reported in
the literature.
For example Eggenberger \cite{Eggenberger97} demonstrated how shape could emerge during growth in response to physical
forces acting on the growing entity.
Bongard \cite{Bongard01} adopted models of genetic regulatory networks to demonstrate how evolution could shape the developmental
trajectories of embodied agents. Later, it was shown how such development could lead to a form of self-scaffolding
that smoothed the fitness landscape and thus increased evolvability \cite{Bongard11}.
Miller \cite{miller2004evolving} introduced a developmental model that enabled growing organisms to regrow structure
removed by damage or other environmental stress.

In the spirit of Beer's minimal cognition experiments \cite{beer1996toward}, we introduce here a minimal model of morphological development in embodied agents (figure \ref{fig:illustrations}).
This model strips away some aspects of other developmental
models, such as those that reorganize the genotype to phenotype mapping
\cite{Eggenberger97, Bongard01, kouvaris2017evolution} or allow the agent's environment to influence its development
\cite{hinton1987learning, miller2004evolving}.
We use soft robots as our model agents since they provide many more degrees of developmental freedom compared to rigid bodies, and can in principle reduce human designer bias.
Here, development is monotonic and irreversible, predetermined by genetic code without any sensory feedback from the environment, and is thus 
\textit{ballistic} in nature rather than adaptive.

While biological development occurs along a time axis, it has been implied 
in some developmental models that time provides only an avenue for regularities to form across space, and that only
the resulting fixed form --- its spatial patterns, repetition and symmetry --- are 
necessary for increasing evolvability.
Compositional pattern producing networks (CPPNs,  \cite{Stanley2007}) explicitly make this assumption in their abstraction of development which collapses the time line to a single point.
While CPPNs have proven to be an invaluable resource in evolutionary robotics \cite{Cheney:2013:UEE:2463372.2463404}, we argue here that discarding time 
may in some cases reduce evolvability
and that there exist fundamental benefits of time itself
in evolving systems.

In this paper, we examine two distinct ways by which ballistic development can increase evolvability.
First, we show how an ontogenetic time scale provides evolution with a 
simple mechanism for inducing mutations with a \textit{range} of magnitude
of phenotypic impact: mutations that occur early in the life time
of an agent have relatively large effects while those that occur
later have smaller effects.
This is important since, according to Fisher's geometric model \cite{fisher1930genetical}, the likelihood a mutation is beneficial is inversely proportional to its magnitude:
Small mutations are less likely to break an \textit{existing} solution.
Larger exploratory mutations, although less likely to be beneficial on average, are more likely to provide an occasional path out of local optima.
Second, we posit that changing ontogenies diversify targets for natural selection to act upon, and that advantageous traits `discovered' by the phenotype during this change can become subject to heritable modification through the `Baldwin Effect' \cite{downing2004development}.

Hinton and Nowlan \cite{hinton1987learning} relied on this second effect when they demonstrated how learning could guide evolution towards a solution to which no evolutionary path led. 
We consider a similar hypothesis with embodied robots and ballistic development, rather than a disembodied bitstring and random search. We demonstrate how open-loop morphological development, without feedback from the environment and without direct communication to the genotype, can similarly alter the search space in which evolution operates making search much easier. 
Hinton \& Nowlan's model of learning was a type of environment-mediated development, in the sense that developmental change stops when the `correct specification' is found, and this information is then used to bias selection towards individuals that find the solution more quickly.
Our work demonstrates that this explicit suppression of development is not necessary; and that completely undirected morphological change is enough to confer evolvability.

\section{Methods}
\label{sec:methods}

All experiments\footnote{\href{https://github.com/skriegman/gecco-2017}{\textbf{https://github.com/skriegman/gecco-2017}} contains the source code necessary for reproducing our results.} were performed in the open-source soft-body physics simulator  
\textit{Voxelyze}, which is described in detail in Hiller and Lipson \cite{hiller2014dynamic}.

\begin{figure}
\hspace{-0.3cm}
\includegraphics[width=0.49\textwidth]{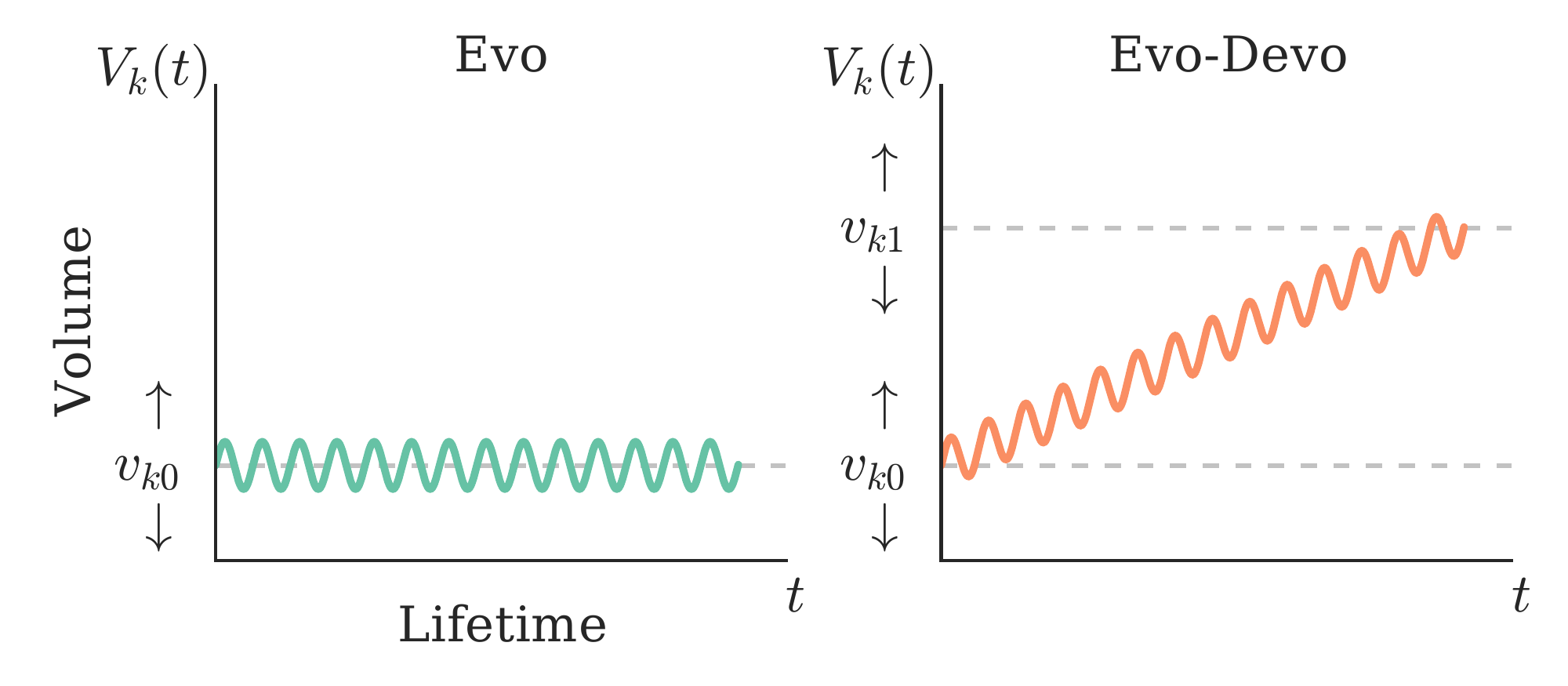}
\vspace{-0.6cm}
\caption{\label{fig:illustrations} \textbf{The voxel picture.} The $k^{\text{th}}$ voxel in an Evo robot maintains a fixed resting volume, $v_{k0}$, throughout the robot's lifetime. Sinusoidal actuation is applied on top of the resting volume. In contrast, the $k^{\text{th}}$ voxel in an Evo-Devo robot changes linearly from a starting volume, $v_{k0}$, to a final volume, $v_{k1}$, over the robot's entire lifetime. Growth, the case when $v_{k1}>v_{k0}$, is pictured here, but shrinkage is also possible and occurs when $v_{k1}<v_{k0}$. 
When $v_{k1}=v_{k0}$, Evo-Devo voxels are behaviorally equivalent to Evo voxels. 
Voxels actuate at 4 Hz in our experiments (for 8 sec or 32 cycles) however actuation is drawn here at a lower frequency to better convey the sinusoidal volumetric structure in time.} 
\end{figure}

We consider a locomotion task for soft robots composed of a $4\times4\times3$ grid of voxels (see figure \ref{fig:videos} for example). 
Each voxel within and across robots is identical with one exception:
its volume.
At any given time, a robot is completely specified by an array of \textit{resting volumes}, one for each of its 48 constituent voxels.
If the resting volumes are static across time then a robot's genotype is this array of 48 voxel volumes; however, because we enforce bilateral symmetry, a genome of half that size is sufficient.
On top of the deformation imposed by the genome, each voxel is volumetrically actuated according to a global signal that varies sinusoidally in volume over time (figure \ref{fig:illustrations}). The actuation is a linear contraction/expansion from their baseline resting volume.

Under this type of rhythmic actuation, many asymmetrical mass distributions will elicit locomotion to some extent. For instance, a simple design, with larger voxels in its front half relative to those in its back half, may be mobile when its voxels are actuated uniformly. Although this design would be rather inefficient since it most likely drags much of its body across the floor as it moves. More productive designs are not so intuitive, even with this fixed controller.

An individual is evaluated for 8 seconds, or 32 actuation cycles.
The fitness was taken to be the distance, in the positive $y$ direction, the robot's center of mass moved in 8 seconds, normalized by the robot's total volume.
Thus, a robot with volume 48 that moves a distance of 48 will have the same fitness --- a fitness of one --- as a similarly shaped but smaller robot with volume 12 that moves a distance of 12. 
Distance here is measured in units that correspond to the length of a voxel with volume one.
If, however, a robot rolls over onto its top layer of voxels it is assigned a fitness of zero and evaluation is terminated. This constraint prevents a rolling ball morphology from dominating more interesting gaits.

We have now built up all of necessary machinery of our first type of robot which we shall call the \textbf{Evo} robot.
Populations of these robots can evolve:
body plans change from generation to generation (phylogeny); but they can not develop: body plans maintain a fixed form, apart from actuation, while they behave within their lifetime (ontogeny).

We consider a second type of robot, the \textbf{Evo-Devo} robot, which inherits all of the properties of the Evo robot but has a special ability: Evo-Devo robots can develop as well as evolve.
These robots are endowed with a \textit{minimally complex} model of development in which resting volumes change linearly in ontogeny. We call this ballistic development to distinguish it from environment-mediated development.
Ballistic development is monotonic with a fixed rate predetermined by a genetic program; its onset and termination are constrained at birth and death, respectively; it is strictly linear, without mid-course correction.
The volume of the $k^{\text{th}}$ voxel in an Evo-Devo robot changes linearly from a starting volume, $v_{k0}$, to a final volume, $v_{k1}$, within the lifetime of a robot (figure \ref{fig:illustrations}). 
Accordingly, the genotype of a robot that can develop is twice as large as that of robots that cannot develop, since there are two parameters ($v_{k0}$ and $v_{k1}$) that determine the volume of the $k^{\text{th}}$ voxel at any particular time. 
Although it is important to note that the space of possible morphologies (collection of resting volumes) is equivalent both with and without development.

\subsection{From gene to volume.}

Like most animals, our robots are bilaterally symmetrical.
We build this constraint into our robots by constraining the 24 voxels on the positive $x$ side of the robot to be equal to their counterparts on the other side of the $y$ axis.
Instead of 48 Evo genes, therefore, we end up with 24.

A single Evo gene stores the resting length, $s_k$, of the $k^{\text{th}}$ voxel, which is cubed to obtain the \textit{resting} volume, $r_k(t)$, at any time, $t$, during the robot's lifetime.
\begin{equation} 
r_k(t)=s_k^3 \qquad k=1,2,\dots,24
\end{equation}
The resting lengths may be any real value in the range $(0.25,\; 1.75)$, inclusive. Note that the resting volume of an Evo robot does not depend on $t$, and is thus constant in ontogenetic time.

Volumetric actuation, $a(t)$ with amplitude $u$, and period $w$, takes the following general form in time.
\begin{equation}
a(t) = u* \sin(2\pi t/w)
\end{equation}
Actuation is limited to $\pm 20\%$ and cycles four times per second ($u=0.20$, $w=0.25$ sec).

However, for smaller resting volumes, the actuation amplitude is limited and approaches zero (no actuation) as the resting volume goes to its lower bound, $0.25^3$. This restriction is enforced to prevent opposite sides of a voxel from penetrating each other, effectively incurring negative volumes, which can lead to simulation instability. 
This damping is applied only where $s_k<1$ (shrinking voxels) and accomplished through the following function.
\begin{equation}
d(s_k) = \begin{cases} 
      1 & s_k \geq 1 \\
      (4s_k-1)/3 & s_k < 1
   \end{cases}
\end{equation}
Thus $d(s_k)$ is zero when $s_k=0.25$, and is linearly increasing in $s_k\le1$. 
The true actuation, $\tilde{a}(t, s_k)$, is calculated by multiplying the unrestricted actuation, $a(t)$, by the limiting factor, $d(s_k)$.
\begin{equation}
\tilde{a}(t,s_k)=a(t)*d(s_k)
\end{equation}

Actuation is then added to the resting volume to realize the \textit{current} volume, $V_k(t)$, of the $k^{\text{th}}$ voxel of an Evo robot at time $t$.
\begin{equation} \label{eq:evo_vol}
V_k(t)=[s_k+\tilde{a}(t,s_k)]^3
\end{equation}

For Evo-Devo robots, a gene is a pair of voxel lengths $(s_{k0}, \; s_{k1})$ corresponding to the $k^{\text{th}}$ voxel's starting and final resting lengths, respectively. Thus, for a voxel in an Evo-Devo robot, the resting volume at time $t \in (0,\tau)$ is calculated as follows. 
\begin{equation} 
r_k(t)=\left[s_{k0} + \frac{t}{\tau} (s_{k1}-s_{k0})\right]^3
\end{equation}
Where the difference in starting and final scale $(s_{k1}-s_{k0})$ determines the slope of linear development which may be positive (growth) or negative (shrinkage). 
The current volume of the $k^{\text{th}}$ voxel of an Evo-Devo robot is then determined by the following.
\begin{equation}
V_k(t)=\left[ r_k^{1/3}(t) + \tilde{a}\left(t, \; r_k^{1/3}(t)\right)\right]^3
\end{equation}
Hence the starting resting volume, $v_{k0}$, and final resting volume, $v_{k1}$, are the current volumes at $t=0$ and $t=\tau$, respectively.
\begin{align}
\begin{split}
v_{k0} &= V_k(0) =  s_{k0} ^3 \\
v_{k1} &= V_k(\tau) = s_{k1} ^3
\end{split}
\end{align}
Note that an Evo gene is a special case of an Evo-Devo gene where $s_{k0}=s_{k1}$, or, equivalently, where $v_{k0}=v_{k1}$.

For convenience, let's define the current \textit{total} volume of the robot across all 48 voxels as $Q(t)$.
\begin{equation}
Q(t)=2\sum_{k=1}^{24} V_k(t)
\end{equation}
We track the $y$ position of the center of mass, $y(t)$, as well as the current total volume, $Q(t)$, at $n$ discrete intervals within the lifetime of a robot. Fitness, $F$, is the sum of the distance traveled in time interval, divided by the average volume in the interval.
\begin{equation} \label{eq:fitness}
F = 2 \sum_{t=1}^{n} \frac{y(t)-y(t-1)}{Q(t) + Q(t-1)}
\end{equation}
We track $y(t)$ and $Q(t)$ 100 times per second. Since robots are evaluated for eight seconds, $n=800$.

\subsection{A direct encoding.}

This paper differs from previous evolutionary robotics work that used Voxelyze \cite{Cheney:2013:UEE:2463372.2463404, cheney2014evolved, Cheney:2015:ESR:2739480.2754662, corucci2016material} in that we evolve the volumes of a fixed collection of voxels, rather than the presence/absence of voxels in a bounding region. 
Another difference is that 
we do not employ the CPPN-NEAT evolutionary algorithm \cite{Stanley2007}, 
but instead use a direct encoding with bilateral symmetry about the $y$ axis.
A comparison of encodings in our scenario is beyond the scope of this paper.
However we noticed that the range of evolved morphologies here, under our particular settings, was much smaller than that of previous work which used voxels as building blocks,
and that it is easier to reach extreme volumes for individual voxels using a direct encoding. 

Apart from the difference in encoding, this work is by in large consistent with this previous work. We use the same physical environment as Cheney et al. \cite{Cheney:2013:UEE:2463372.2463404}: a wide-open flat plain. The material properties of our voxels are also consistent with the `muscle' voxel type from the palette in this work;  
although these voxels had a fixed resting volume of one ($s_k=1$ for all $k$).
Our developmental mechanism is strongly based on Corucci et al. \cite{corucci2016material}, which used volumetric deformation in a closed-loop pointing task.

\subsection{Evolutionary search.}

We employ a standard evolutionary algorithm, Age-Fitness-Pareto Optimization (AFPO, \cite{Schmidt2011}), which uses the concept of Pareto dominance and an objective of age (in addition to fitness) intended to promote diversity among candidate designs. For 30 runs, a population of 30 robots is evolved for 2000 generations.
Every generation, the population is first doubled by creating modified copies of each individual in the population. Next, an additional random individual is injected into the population. Finally, selection reduces the population down to its original size according to the two objectives of fitness (maximized) and age (minimized).

The same number of parent voxels are mutated, on average, in both Evo and Evo-Devo children.
Mutations follow a normal distribution ($\sigma=0.75$) and are applied by first choosing what parameter types to mutate, and then choosing which voxels to mutate.
For Evo robots, we simply visit each voxel (on the positive $x$ side) of the parent and, with probability 0.5, mutate its single parameter value.
For Evo-Devo parents, we flip a coin for each parameter to be mutated (if neither will be mutated, flip a final coin to choose one or the other).  
This results in a 25\% chance of mutating both, and a 37.5\% chance of mutating each of the two individual parameters alone.
Then we apply the same mutation process as before in Evo robots: loop through each voxel of the parent and, with probability 0.5, mutate the selected parameter(s).

\subsection{An artificially rugged landscape.}

We did not fine-tune the mutation hyperparameters (scale and probability), but intentionally chose a relatively high probability of mutation in order to elicit a large mutational impact in an attempt to render evolutionary search more difficult. 
This removes easy to follow gradients in the search space --- `compressing' gentle slopes into abrupt cliffs --- which make `good designs' more difficult to find.  Any one of these good solutions then, to a certain extent, become like Hinton \& Nowlan's `needle in a haystack' \cite{hinton1987learning}.

Note that there are other ways to enforce rugged fitness landscapes, and such landscapes are naturally occurring in many systems, though our particular task/environment is not one of them. Future work should investigate these tasks and environments with a fine-tuned mutation rate.

\section{Results}
\label{sec:results}

In this section we present the results of our experiments\footnote{\href{https://youtu.be/gXf2Chu4L9A}{\textbf{https://youtu.be/gXf2Chu4L9A}} directs to a video overview of our experiments.} and indicate statistical significance under the Mann-Whitney $U$ test where applicable.

\subsection{Random search.}

\begin{figure}
\includegraphics[width=0.5\textwidth]{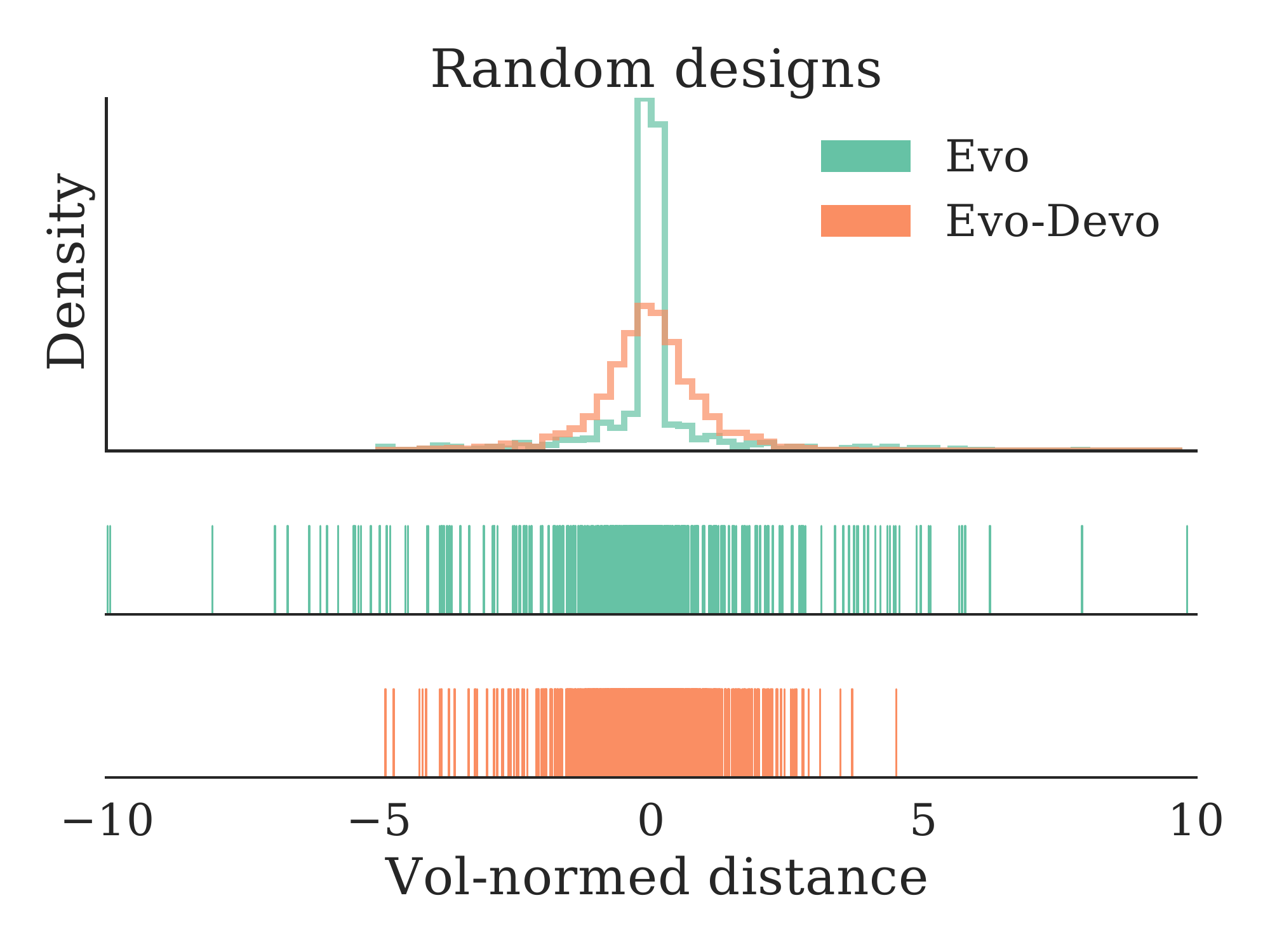}
\vspace{-0.6cm}
\caption{\label{fig:random} One thousand randomly generated robots for each group. 
The horizontal axes measure fitness: volume normalized distance in the positive $y$ direction.
The best overall designs are the best Evo robots since they maintain their good form as they behave. However, most designs are immobile (mode at zero) and Evo-Devo robots are more likely to move (less mass around zero) since they explore a continuum of body plans rather than a single static guess.}
\end{figure}

To get a sense of the evolutionary search space, prior to optimization, we randomly generated one thousand robots from each group (figure \ref{fig:random}).
The horizontal axes of figure \ref{fig:random} measure the fitness (equation \ref{eq:fitness}) of our randomly generated designs.
The top portion of this figure plots the histogram of relative frequencies, using equal bin sizes between groups.
The mode is zero for both groups, meaning that the majority of designs are immobile.

The best possibility here is to randomly guess a good Evo robot since this good morphology is utilized for the full 32 actuation cycles. 
This is why the best random designs are Evo robots.
However, the Evo-Devo distribution contains much less mass around zero than the Evo distribution. It follows that it is more likely that an Evo-Devo robot moves at all, if only temporarily, since this only requires some interval of the many morphologies it sweeps over to be mobile. 
Also note that while the total displacement may be lower in the Evo-Devo case, since these robots `travel' through a number of different morphologies, they may pass through those which run at a higher instantaneous speed (but spend less of their lifetime in this morphology).

\subsection{Evolution.}

\begin{figure}
\hspace{-0.2cm}
\includegraphics[width=0.48\textwidth]{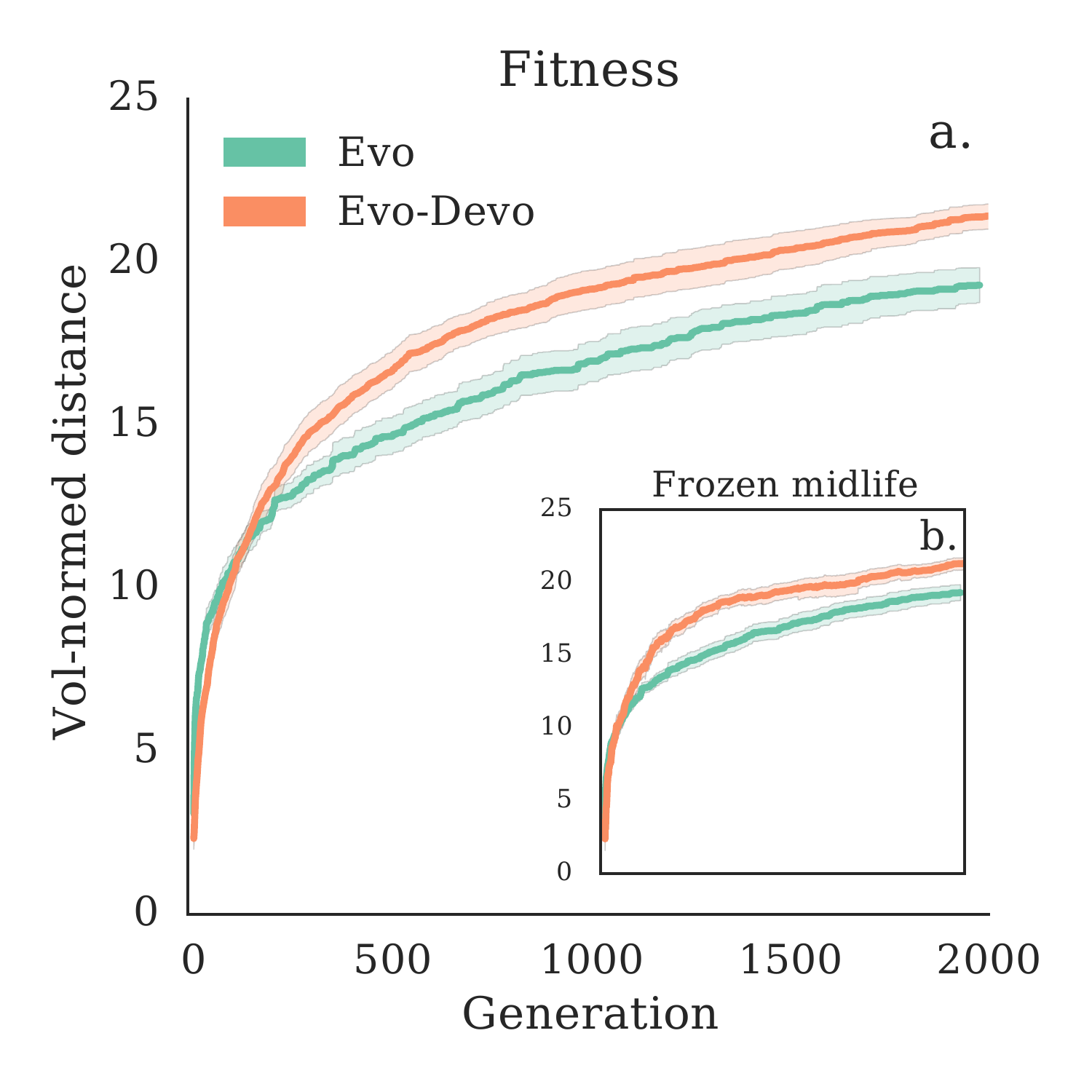}
\vspace{-0.7cm}
\caption{\label{fig:main_frozen} For thirty runs, a population of thirty robots is evolved for two thousand generations. (a) Best of generation fitness for Evo and Evo-Devo robots. (b) The same robots are reevaluated with development frozen at their midlife morphology.
Means are plotted with 95\% bootstrapped confidence intervals.}
\end{figure}

The results of the evolutionary algorithm are displayed in figure \ref{fig:main_frozen}a. 
In the earliest generations, evolution is consistent with random search and the best Evo robots start off slightly better than the best Evo-Devo robots.
However, the best Evo-Devo robots quickly overtake the best Evo robots. 
At the end of optimization there is a significant difference between Evo and Evo-Devo run champions ($U=122, \; p<0.001$).

We also chose to reevaluate Evo-Devo robots with their development frozen at their median ontogenetic morphologies (figure \ref{fig:main_frozen}b). For each robot, we measure the robot's fitness (equation \ref{eq:fitness}) at this midlife morphology with development frozen, for two seconds. Selection is completely blind to this frozen evaluation. It exists solely for the purpose of post-evolution analysis, and serves primarily as a sanity check to make sure Evo-Devo robots are not explicitly utilizing their ability to grow/shrink to move faster.

Development appears to inhibit locomotion to some degree as the best morphologies run slightly faster with development turned off, particularly in earlier generations.
A significant difference, at the 0.001 level, between Evo robots and Evo-Devo robots with development frozen at midlife, occurs after only 108 generations compared to 255 generations with development enabled.
Note that the midlife morphology is not necessarily the top speed of an Evo-Devo robot. In fact it is almost certainly not the optimal ontogenetic form since the best body plan may occur at any point in its continuous ontogeny, including the start and endpoints.

\subsection{Closing the window.}

Once an Evo-Devo robot identifies a good body plan in its ontogenetic sweep, its descendants can gain fitness by `suppressing' development around the good plan through heterochronic mutations.
This can be accomplished by incrementally closing the \textit{developmental window}, the interval $(s_{k0},\; s_{k1})$, for each voxel, around the good morphology.
In the limit, under a fixed environment, this process ends with a decedent born with the good design from the start and devoid of any developmental at all ($s_{k0} = s_{k1}$ for all voxels).
This phenomenon, best known as the Baldwin Effect, is instrumental in evolution
because natural selection is a hill-climbing process and therefore blind to needles in a haystack, good designs (local optima) to which no gradient of increased fitness leads.
The developmental sweep, however, alters the search space in which evolution operates, surrounding the good design
by a slope which natural selection can climb \cite{hinton1987learning}.

To investigate the relationship between development and fitness, we add up all of the voxel-level development windows to form a individual-level summary statistic, $W$. We define the \textit{total} development window, $W$, as the sum of the absolute difference of starting and final resting lengths across the robot's 48 voxels. 
\begin{equation}
W = \sum_{k=1}^{48} \text{abs}(s_{k1}-s_{k0})
\end{equation}

\begin{figure}[t]
\hspace{-0.2cm}
\includegraphics[width=0.48\textwidth]{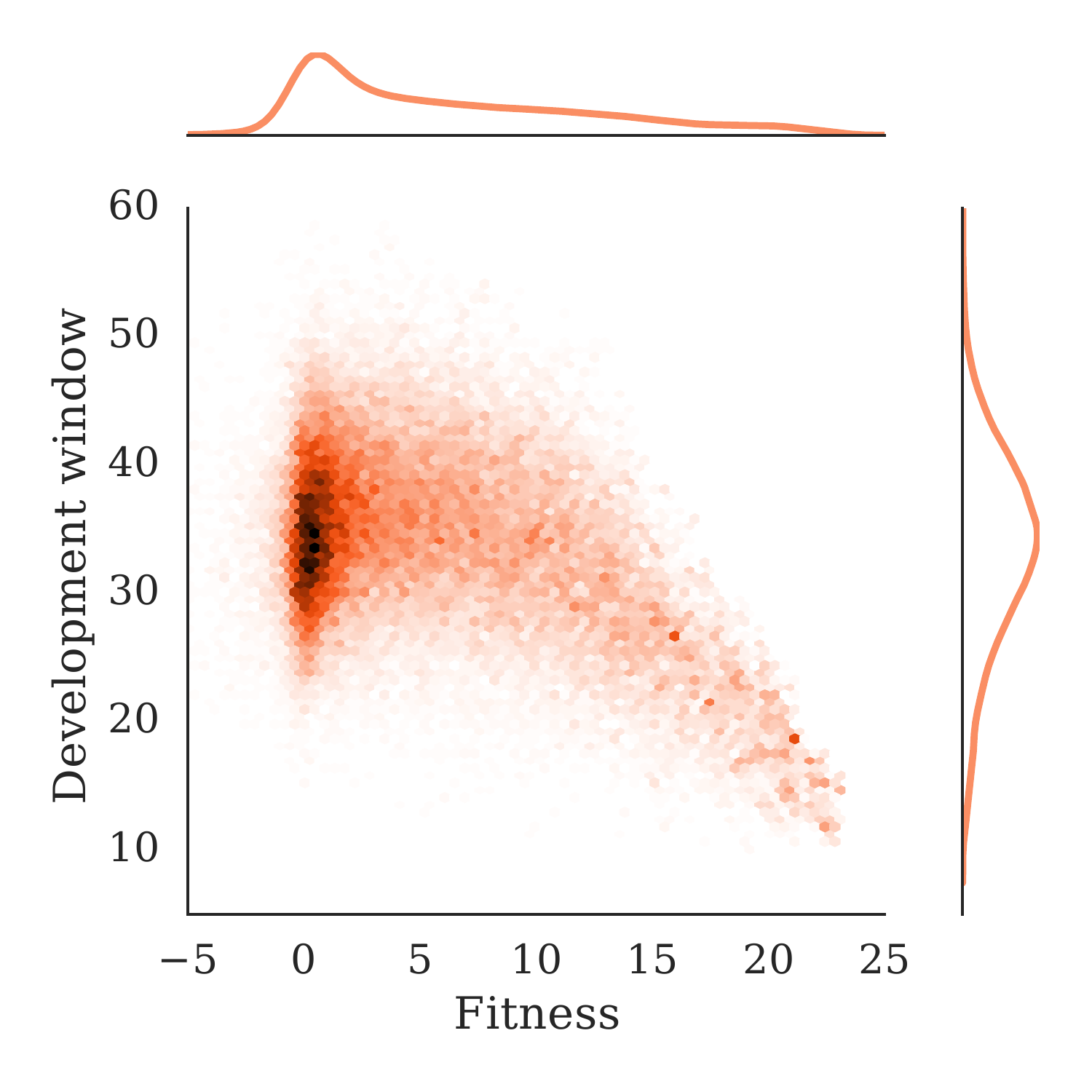}
\vspace{-0.7cm}
\caption{\label{fig:window} The relationship between the amount of development at the individual level ($W$) and fitness ($F$). The fastest individuals have small developmental windows surrounding a fast body plan.}
\end{figure}

\begin{figure}
\hspace{-0.2cm}
\includegraphics[width=0.48\textwidth]{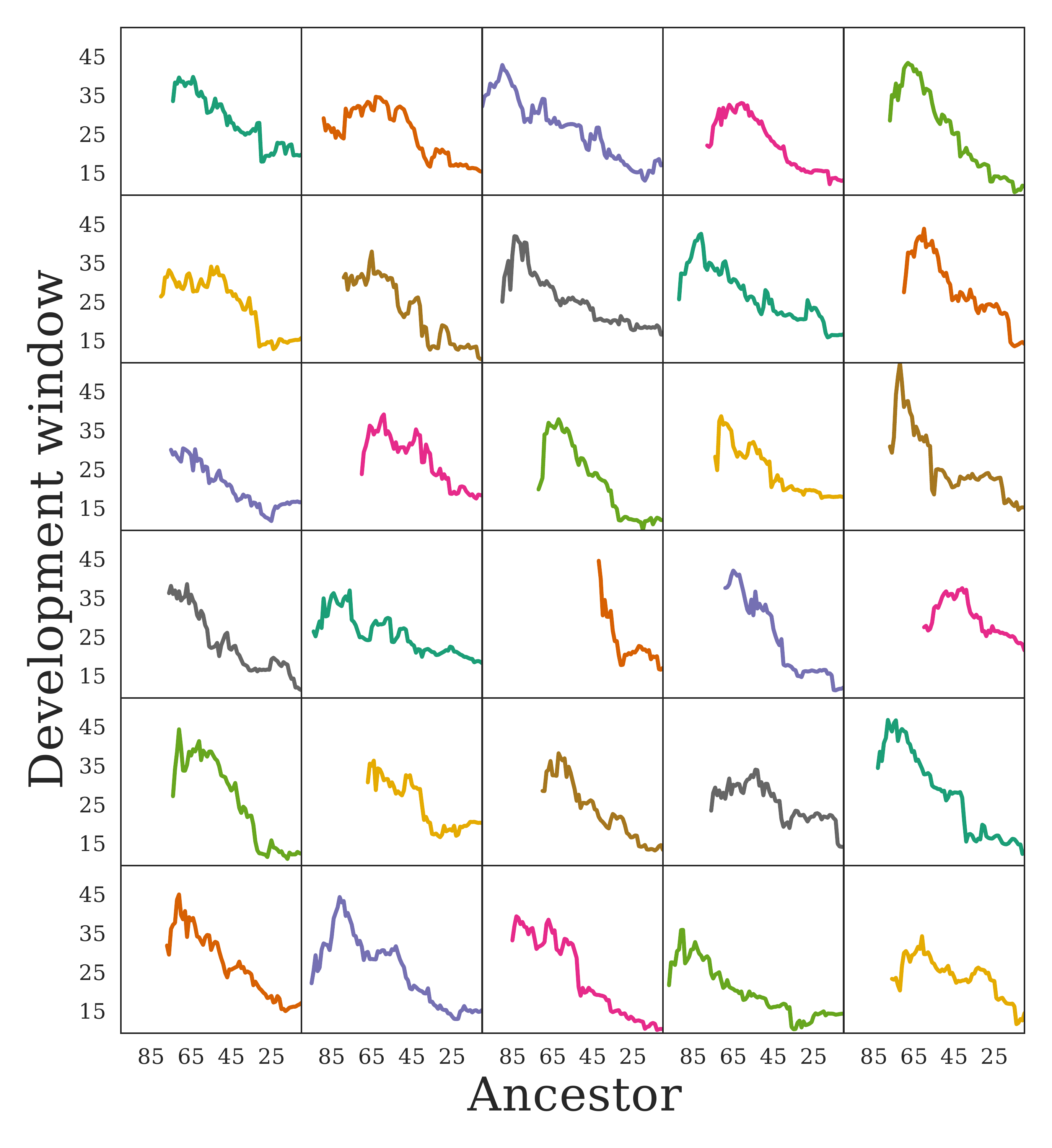}
\vspace{-0.7cm}
\caption{\label{fig:ancestral_window} Closing the window. Total development window trajectories (in phylogeny) of the lineages of the most fit individuals in each run. Phylogenetic time goes from left to right: from the oldest ancestor (randomly created) to its most recent decedent, the current run champion.}
\end{figure}

Overall there is a strong negative correlation between fitness, $F$, and the total development window, $W$, in Evo-Devo robots (figure \ref{fig:window}). 
To achieve the highest fitness values a robot needs to have narrow developmental windows at the voxel level.
However, this statistic doesn't discriminate between open/closed windows early/late in evolution. 
To show what sorts of development window/evolutionary time relationships eventually lead to highly fit individuals, we grab the lineages of only the most fit individuals at the end of evolutionary time (figure \ref{fig:ancestral_window}). 
In the most fit individuals, development windows tend to first increase slightly in phylogeny before decreasing to their minimum, or close nearby.
The age objective in AFPO lowers the selection pressure on younger individuals which allows them to explore, through larger developmental windows, a larger portion of design space until someone in the population discovers a locally optimal solution which creates a new selection pressure for descendants with older genetic material to `lock in' or canalize this form with smaller developmental windows.
These results further suggests that development itself is not optimal,
it is only helpful in that it can lead to better optima down the road once the window is closed.

\subsection{The effect of mutations.}

\begin{figure*}
\includegraphics[width=0.495\textwidth]{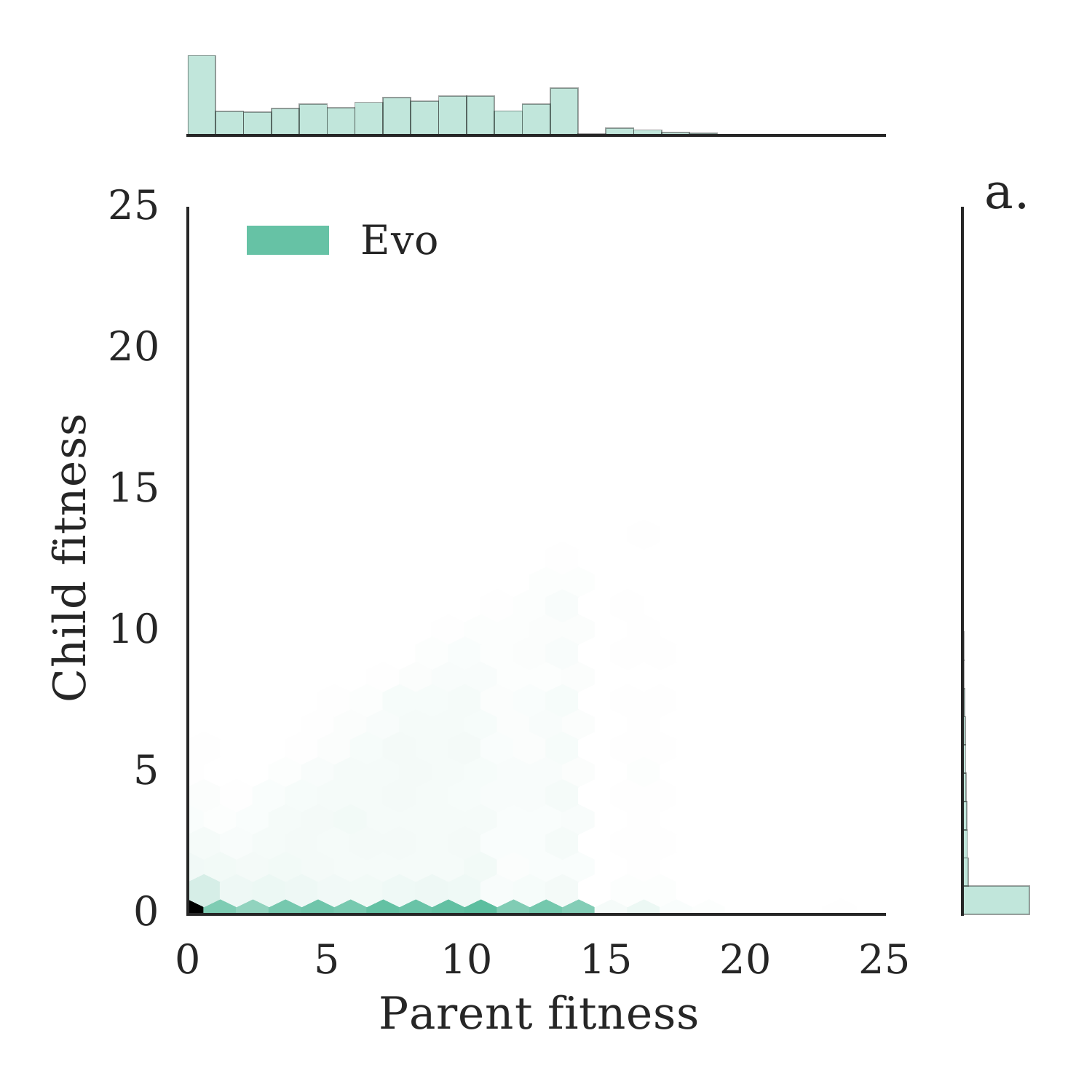}
\includegraphics[width=0.495\textwidth]{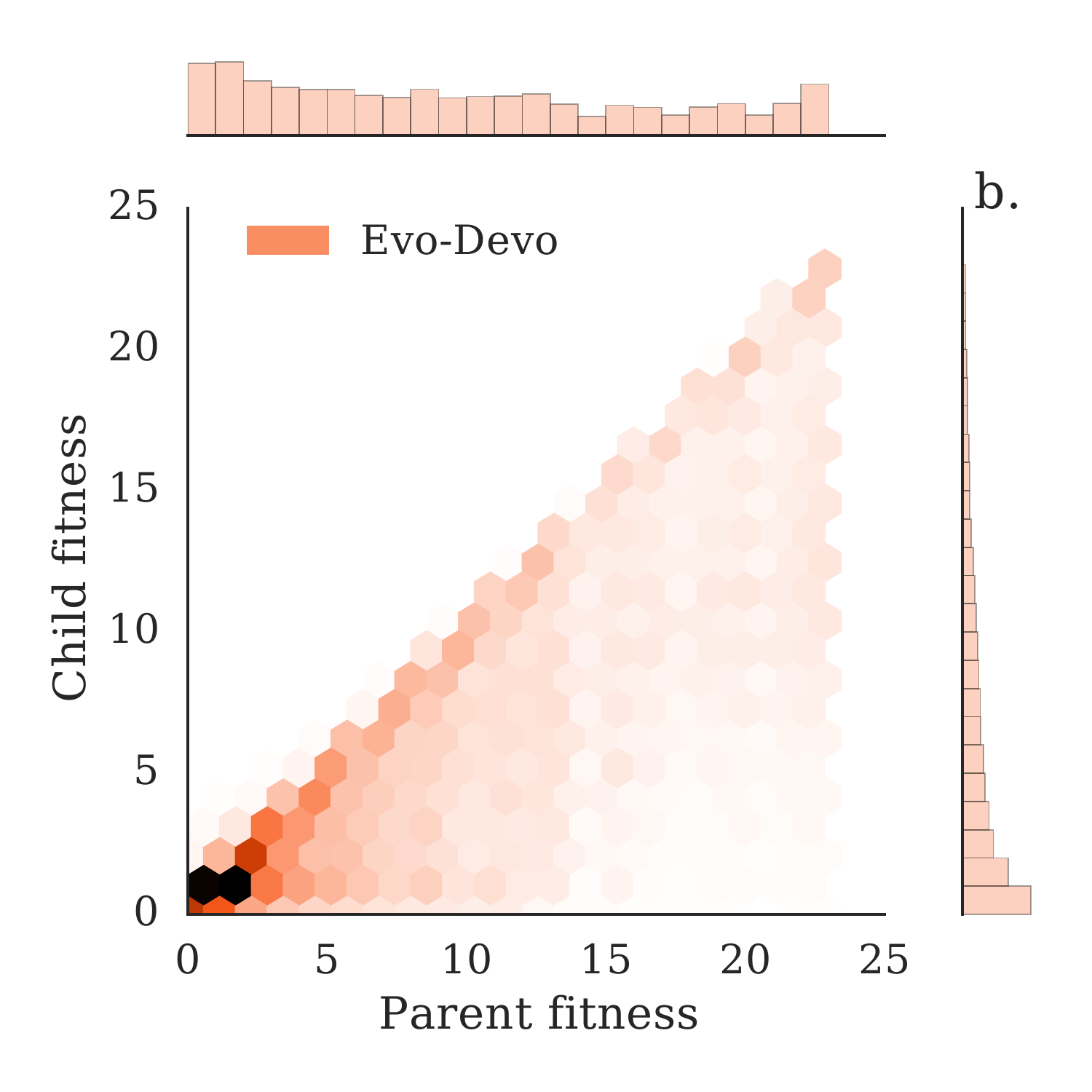}
\vspace{-0.5cm}
\caption{\label{fig:mutations} Mutation impact: child fitness by parent fitness (vol-normed distance). The diagonal represents a neutral mutation, equivalent child and parent fitness. Hexagon bins below the diagonal represent detrimental mutations (child less fit than its parent); bins above the diagonal represent beneficial mutations (child more fit than its parent).}
\end{figure*}

In addition to the parameter-sweeping nature of its search,
developmental time provides evolution with a 
simple mechanism for inducing mutations with a range of magnitude
of phenotypic impact.
The overall mutation impact in our experiments is conveyed in figure \ref{fig:mutations} through 2D histograms of child and parent fitness. Recall that a child is created through mutation by each individual (parent) in the current population. These plots include the entire evolutionary history of all robots in every run. There are relatively so few robots with negative fitness that the histograms need not extend into this region since they contain practically zero density and would appear completely white.

The diagonal represents equal parent and child fitness, a behaviorally neutral mutation. Hexagons below the diagonal represent detrimental mutations: lower child fitness relative to that of its parent. Hexagons above the diagonal represent beneficial mutations: higher child fitness relative to that of its parent.
Mutations are generally detrimental for both groups, particularly in later generations once evolution has found a working solution.
For Evo robots (figure \ref{fig:mutations}a),
most if not all of the mass in the marginal density of child speed is concentrated around zero.
This means that mutations to an Evo robot are almost certain to break the existing parent solution, rendering a previously mobile design immobile.

The majority of Evo-Devo children, however, are generally concentrated on, or just below the diagonal in figure \ref{fig:mutations}b. 
This general pattern holds even in later generations when evolution has found working solutions with high fitness.
It follows that mutations to an Evo-Devo robot may be phenotypically smaller than mutations to an Evo robot, even though they use the same mutation operator. 
Furthermore, figure \ref{fig:mutations}b displays a high frequency of mutations with a wide range of magnitude of phenotypic impact including smaller, low-risk mutations which are useful for refining mobile designs; as well as a range of larger, higher-risk mutations which occasionally provide the high-reward of jumping into the neighborhood of a more fit local optima at a range of distances in the fitness landscape.

Now let's define the impact of developmental mutations, $M$, as the relative difference in child ($F_C$) and parent fitnesses ($F_P$), for positive fitnesses only.
\begin{equation}
M = \frac{F_C }{F_P}-1; \qquad F_C, \; F_P>0
\end{equation}
Then the average mutational impact for early-in-the-life mutations (any mutations that, at least in part, modify initial volumes) is $M_0=-0.29$.
While the average mutational impact for late-in-the-life mutations (that modify final volumes) is $M_1=-0.10$.
Although both types of mutations are detrimental on average, later-in-life mutations are more beneficial (less detrimental) on average ($p<0.001$).
This makes sense in a task with dependent time steps since a child created through a late-in-life mutation will at least start out with the same behavior as its parent and then slowly diverge over its life. Whereas an early-in-life mutation creates a behavioral change at $t=0$.

\subsection{The necessity of development.}

\begin{figure}
\hspace{-0.2cm}
\includegraphics[width=0.48\textwidth]{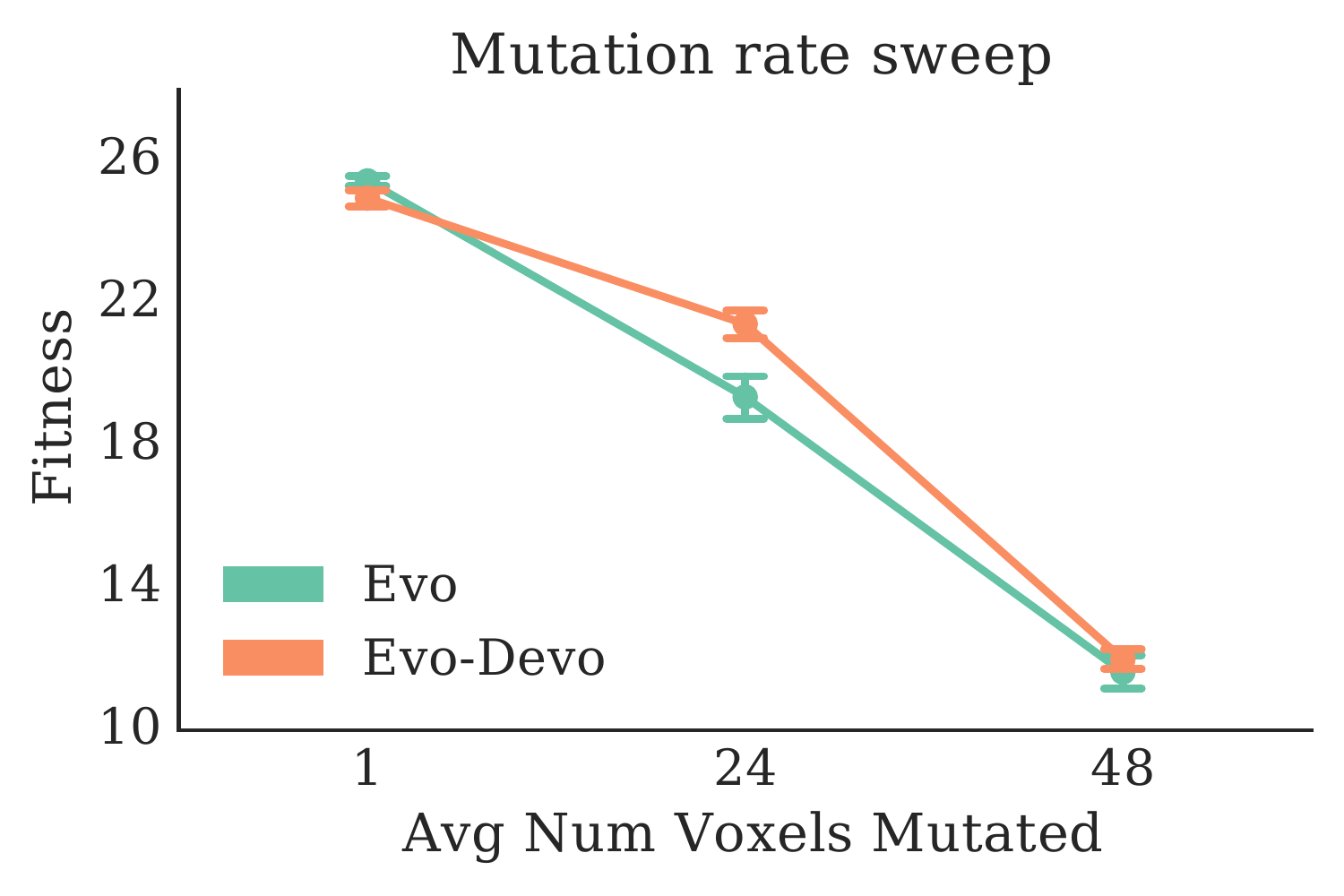}
\vspace{-0.7cm}
\caption{\label{fig:hyperparam_sweep} A hyperparameter sweep of mutation rate: a probability dictating the average number of voxels mutated in an individual robot.}
\end{figure}

In attempting to induce a needle-in-the-haystack fitness landscape, as a proof of concept, we intentionally set the mutation rate and scale fairly high.
A low-resolution hyperparameter sweep (figure \ref{fig:hyperparam_sweep}) indicates that the efficacy of ballistic development is indeed dependent on the mutation rate: there is no significant difference between Evo and Evo-Devo at either very low or very high rates. 
Higher fitness values are obtained through smaller mutation rates, which raises the question: Is development useful only in its ability to decrease the phenotypic impact of mutations?
If so we might prefer Evo robots (with a low mutation rate) since they reside in a smaller search space. 
But how low should the mutation rate be? 
It may in fact be difficult to know \textit{a priori} which mutation rate is optimal.
It is also important to recognize that while we use mutation rate here to artificially tune the ruggedness of the fitness landscape, in a naturally rugged landscape we presumably would not have direct access to such an easily tunable parameter to `undo', or smooth-out the ruggedness.

Moreover, we know that there exist contexts in which developmental flexibility can permit the local speeding up of the basic, slow process of natural selection, thanks to the Baldwin Effect \cite{dennett2003baldwin}. 
Our new data suggests that even open-loop morphological change increases the probability of randomly finding (and subsequently `locking in') a mobile design (figure \ref{fig:random}), and that this probability is increasing in the amount of change (figure \ref{fig:ancestral_window}) even though ballistic development and fitness are inversely correlated (figure \ref{fig:window}). 
The staticity of Evo robots prevents this local speed-up which can place them at a significant disadvantage in rugged fitness landscapes.

\section{Conclusion}

In this paper we introduced a minimal yet embodied model of development in order to isolate the intrinsic effect of morphological change in ontogenetic time,
without the confounding effects of environmental mediation.
Even our simple developmental model naturally provides
a continuum in terms of the magnitude of mutational phenotypic impact,
from the very large (caused by early-in-life developmental mutations) 
to the very small (caused by late-in-life mutations). We predict that,
because of this, such a developmental system will be more evolvable
than an equivalent non-developmental system because the latter lacks
this inherent spectrum in the magnitude of mutational impacts.

We showed that even without any sensory feedback, open-loop development can confer evolvability because it allows evolution to sweep over a much larger range of body plans. Our results suggest that widening the span of the developmental sweep increases the likelihood of stumbling across locally optimal designs otherwise invisible to natural selection, which automatically creates a new selection pressure to canalize development around this good form.
This implies that species with completely blind developmental plasticity tend to evolve faster and more `clearsightedly' than those without it.

Future work will involve closing the developmental feedback loop with as little additional machinery as possible to determine when and how such added complexity increases evolvability.

\section{Acknowledgements}

We would like to acknowledge financial support from
NSF awards PECASE-0953837 and INSPIRE-1344227
as well as the Army Research Office contract W911NF-16-1-0304.
N. Cheney is supported by NASA Space Technology Research Fellowship \#NNX13AL37H.
F. Corucci is supported by grant agreement \#604102 (Human Brain Project) funded by the European Union Seventh Framework Programme (FP7/2007-2013).
We also acknowledge computation provided by the Vermont Advanced Computing Core.

\bibliographystyle{ACM-Reference-Format}
\bibliography{2017_GECCO_Kriegman} 

\end{document}